# Robotic Assembly Control Reconfiguration Based on Transfer Reinforcement Learning for Objects with Different Geometric Features

Yuhang Gai, Bing Wang, Jiwen Zhang, Dan Wu*, and Ken Chen
All authors are from the State Key Laboratory of Tribology in Advanced Equipment,
Department of Mechanical Engineering, Tsinghua University, Beijing, China.
*Corresponding author: Dan Wu (phone: 1-391-083-2965; e-mail: wud@mail.tsinghua.edu.cn).

**Abstract**—Robotic force-based compliance control is a preferred approach to achieve high-precision assembly tasks. When the geometric features of assembly objects are asymmetric or irregular, reinforcement learning (RL) agents are gradually incorporated into the compliance controller to adapt to complex force-pose mapping which is hard to model analytically. Since force-pose mapping is strongly dependent on geometric features, a compliance controller is only optimal for current geometric features. To reduce the learning cost of assembly objects with different geometric features, this paper is devoted to answering how to reconfigure existing controllers for new assembly objects with different geometric features. In this paper, model-based parameters are first reconfigured based on the proposed Equivalent Theory of Compliance Law (ETCL). Then the RL agent is transferred based on the proposed Weighted Dimensional Policy Distillation (WDPD) method. The experiment results demonstrate that the control reconfiguration method costs less time and achieves better control performance, which confirms the validity of proposed methods.

## 1. Introduction

### 1.1. Robotic Assembly Control

Traditional robotic off-line programming technology cannot adapt to assembly tasks in dynamic environments and also cannot ensure the precision of assembly tasks [1]. Hence, robots need to be guided by external feedback of vision or force information to execute assembly tasks [2]. Since the precision of visual signals is easily affected by the environment, the assembly performance with vision feedback is not as optimal as that with force feedback in most cases.

Force-guided control methods are also usually known as compliance control methods. According to whether the control method depends on model information, compliance control methods can be categorized into model-based methods, model-free methods, and model-hybrid compliance control (MHCC) methods [3].

Model-based compliance control methods are dedicated to describing the relationship between detected force/moment and the relative pose of assembly objects through force-pose mapping. According to force feedback, the robot will correct the assembly path to avoid collision and jamming between assembly objects. According to whether the structure and parameters of the compliance controller are adaptive during the assembly task, model-based compliance control can be divided into constant compliance control [4] - [7] and adaptive compliance control methods [8] - [10]. Because the force-pose mapping of the assembly process is always nonlinear and time-variant, constant compliance control methods cannot guarantee the stability and rapidity of the control process at the same time. Hence, adaptive compliance control methods are proposed to improve control performance [8] - [12]. However, adaptive compliance control methods still rely on modeling and human experience. As modeling the assembly process of asymmetric objects is very difficult, model-based approaches can only be based on a simplified model and cannot achieve high control performance.

Model-free compliance control methods mainly belong to data-driven learning methods [13] - [17]. Model-free compliance control methods are capable to solve assembly tasks of asymmetric objects because they do not rely on the analytical model of the assembly process. [13] proposes a compliance control method based on Gauss mixture model, which is regressed from human demonstration data. Then the inverse kinematic of Gauss mixture model replaces the traditional analytical compliance law to control the assembly process. The performance of compliance control based on Gauss mixture model is limited by the quality of teaching data. Hence, RL methods are widely employed to improve the optimality of controllers. However, RL suffers from low sampling efficiency, even with plenty of optimization mechanisms, e.g. fuzzy reward [15], hierarchical RL [16], dimension extension [17], etc.

MHCC methods combine model-based and model-free methods to improve the sampling efficiency under the premise of ensuring the optimality of controllers [15] - [19]. In practical applications, assembly tasks of asymmetric objects require not only control precision but also the security and rapidity of the training process. Hence, MHCC is more widely employed to perform assembly tasks of asymmetric objects. Residual RL belongs to MHCC method [18], [19], in which the action of an RL agent and the output of a constant compliance controller are added to command the robot. In MHCC method in [15] - [17], the learned part is used to adaptively coordinate the parameters of the model-based part according to the feedback signal. In our paper, we employ MHCC method like [15] - [17] to construct the controller. Since the structure of MHCC controllers is more complicated, the reconfiguration method of MHCC controllers is worth exploring.

### 1.2. Control Reconfiguration and Transfer RL

Control reconfiguration is an approach to adapting to the change in plant dynamic by adjusting the structure and parameters of the controller [20]. When the geometric features of assembly objects change, the plant dynamic of the assembly process will be different. To ensure the successful execution of the assembly task, the structure and parameters of the controller should be reconfigured [21] - [26]. In this paper, we implement MHCC method, which combines an RL agent and some model-based modules to solve assembly tasks with different geometric features. Because model-based parameters and the RL agent are coupled in assembly tasks, the accuracy of model-based reconfiguration affects the complexity of RL agent transfer. Specifically, if model-based reconfiguration is very inaccurate, RL agents in source and target domains may be extremely different. Then the RL agent in the source domain cannot teach, but rather misdirect the RL agent in the target domain. Hence, the crucial issues of reconfiguration of an MHCC controller are:

1. Reconfigure model-based parameters as accurately as possible to reduce the gap between source and

target domains.

2. Learn more detailed knowledge from more similar source domains.

The reconfiguration of model-based parameters needs a fault diagnosis process to describe the change in plant dynamic, then the control reconfiguration process is implemented based on fault diagnosis [27] - [31]. The key lies in constructing the relationship between the geometric features of assembly objects and control parameters. Since the modeling and fault diagnosis process cannot cover all the factors of non-linear and coupling force-pose mapping, model-based reconfiguration cannot solve new assembly tasks well alone. But model-based reconfiguration can reduce the gap between the RL agents in existing assembly tasks and new assembly tasks, which is meaningful for RL transfer.

The reconfiguration of RL agents belongs to the field of transfer learning [32] - [36]. RL Transfer methods usually influence the training process in the target domain by modifying reward function or loss function in the target domain, which are known as reward shaping methods and policy distillation methods [37], [38]. Policy distillation methods extract a basic policy from all existing policies in source domains. Then the training process in the target domain can be guided towards the basic policy by taking the term of learning from the basic policy as part of loss function [39] - [40]. [41] further confirms that the policy in the target domain should not only learn from the basic policy but also explore to ensure optimality. However, there are still dilemmas in existing RL transfer methods on assembly tasks. The most realistic one is that the transfer may be negative when the task in the source domain is quite different from the task in the target domain. In our paper, we try to optimize policy distillation through similarity weight and dimensional transfer mechanisms.

### 1.3. Motivation and Contribution

Motivated by solving assembly tasks of asymmetric objects with different geometric features, this paper proposes a control reconfiguration method for MHCC controllers. The control reconfiguration process is divided into model-based reconfiguration and RL agent transfer in sequence. Firstly, this paper proposes ETCL to relate model-based parameters with the geometric features of assembly objects. Then this paper proposes WDPD method to transfer the RL agent under the guidance of dimensional similarity weight.

The rest of the paper is organized as follows. Section 2 gives a brief introduction about MHCC controllers in assembly tasks. Section 3 develops the control reconfiguration method of MHCC controllers. Section 4 provides experiment verifications. Section 5 summarizes the work of this paper.

### 2. Control Method

The basic idea of MHCC method is to first construct control modules by employing as much model information as possible, and then to train an RL agent to modify the parameters in modules adaptively. The total control framework and design process of the MHCC controller are shown in Fig. 1. The geometric features of assembly objects are known in advance. State and output equations of the assembly process are built based on the known geometric features of assembly objects. Then corresponding model-based control modules are designed. As indicated in [42], significant nonlinear and coupling characteristics exist in the assembly process, which cannot be well settled by constant model-based control modules. Then, an RL agent is employed adaptively adjust the model-based control modules to enhance the control performance.

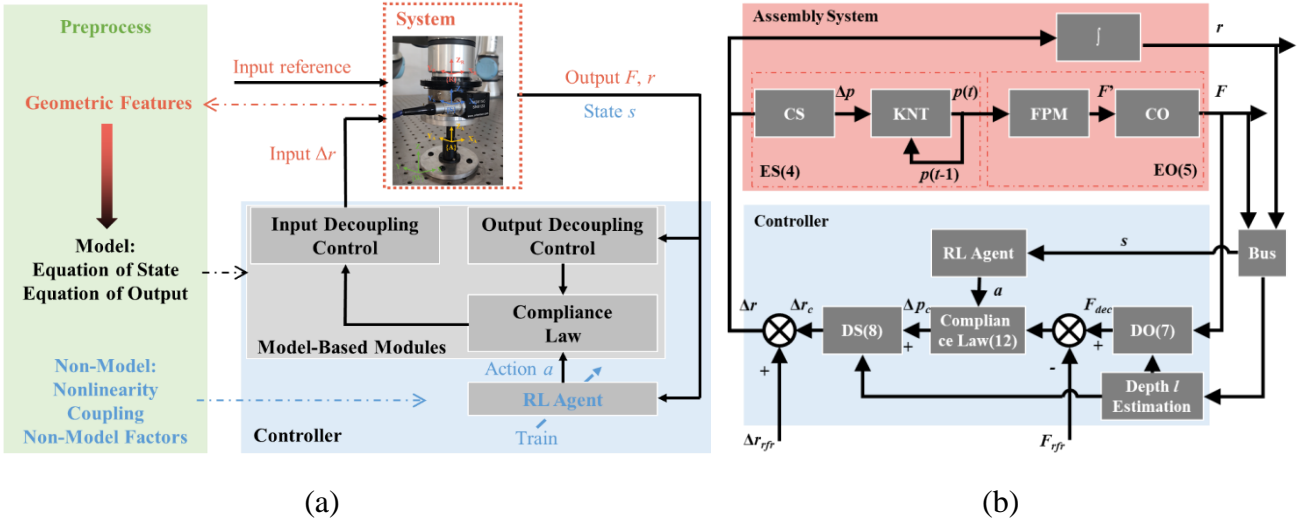

(a)             (b)

Fig. 1. (a) Design process of MHCC controllers; (b) Control block diagram in assembly tasks.

## 2.1. Model of Assembly process

We focus on the assembly tasks of columnar objects with equal cross section. The axis of the peg is composed of all gravity centers of all cross sections. As shown in Fig. 2, coordinate systems are defined to assist in describing the assembly process. The Cartesian coordinate system is named as the world coordinate system {W}. When the peg is inserted into the hole with length $l$, the assembly coordinate system {A} is defined at the midpoint of the inserted part of the peg axis. The direction Z of {A} is along the axis of the hole. {A} is dynamic as the peg is inserted continuously. The force-moment sensor coordinate system {S} and robot coordinate system {R} are also necessary in a force-guided controller. {S} is fixed on the center of force-moment sensor, in which the output of the assembly process is collected. {R} is fixed on the TCP of the robot, in which the input of the assembly process is provided. {S}, {R}, and {A} are not coincident in {W}.

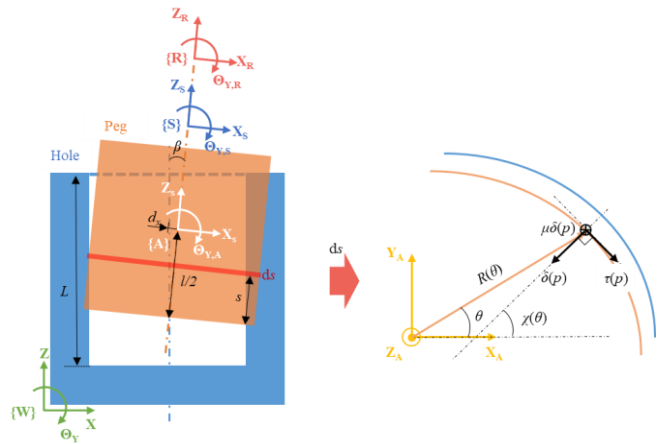

Fig. 2. Building force-pose mapping of the assembly process through the calculus method. Integral variables are length $s$ and angle $\theta$. The contact area is first sliced along the axis of the peg and then each slice is divided along angle $\theta$.

As indicated in [42], the assembly process is modelled as a MIMO system, where the relative pose between the peg and hole is defined as the state of the system.

$$p = [d_x, d_y, l, \alpha, \beta, \gamma], d_x, d_y, l, \alpha, \beta, \gamma \in \mathbb{R} \tag{1}$$

where $\alpha$, $\beta$, $\gamma$ are the RPY angle of the peg in {A}. $l$ is the inserted length of the peg. $d_x$ and $d_y$ are the bias between the peg axis and the hole axis at midpoint of the insertion depth. The output of the system includes force/moment information $F$ and Cartesian pose of the robot $r$.

$$\begin{aligned} F &= [F_x, F_y, F_z, M_x, M_y, M_z] \\ r &= [x, y, z, \theta_x, \theta_y, \theta_z] \end{aligned} \tag{2}$$

where $F_x$, $F_y$, $F_z$ are forces along axes of {W}. $M_x$, $M_y$, $M_z$ are moments along axes of {W}. $x$, $y$, $z$ are the position of {R} in {W}, $\theta_x$, $\theta_y$, $\theta_z$ are the orientation of {R} in {W}. The incremental motion of the robot is defined as the input of the system.

$$\Delta r = [\Delta x, \Delta y, \Delta z, \Delta \theta_x, \Delta \theta_y, \Delta \theta_z] \tag{3}$$

where $\Delta$ indicates incremental motion of a variable.

State equation describes the change of relative pose between the peg and hole caused by incremental motion of the robot.

$$\begin{aligned} p(t+1) &= \mathrm{ES}(p(t), \Delta r(t) | R_R^A, t_R^A) \\ p(t+1) &= \mathrm{KNT}(p(t), \Delta p(t)) \\ \Delta p(t) &= \mathrm{CS}(\Delta r(t) | R_R^A, t_R^A) \end{aligned} \tag{4}$$

where ES( ) represents state equation, which is composed of two parts: kinematic function KNT( ) as well as coupling caused by misalignment between {A} and {R} CS( ). $R_R^A$ and $t_R^A$ represent the rotation matrix and translation vector of {A} in {R}. The details about state equation can be referred to [42].

Output equation on force/moment part describes the force/moment collected by sensor caused by the state.

$$\begin{aligned} F(t) &= \mathrm{EO}(p(t) | R_S^A, t_S^A) \\ F'(t) &= \mathrm{FPM}(p(t)) \\ F(t) &= \mathrm{CO}(F'(t) | R_S^A, t_S^A) \end{aligned} \tag{5}$$

where $F'$ represents the force/moment in {A}. EO( ) represents output equation, which is composed of two parts: force-pose mapping FPM( ) as well as coupling caused by misalignment between coordinates systems {A} and {S} CO( ). $R_S^A$ and $t_S^A$ represent the rotation matrix and translation vector of {A} in {S}. The details about output equation can be referred to [12] and [42].

As shown in Fig. 2, when state $p$ is known, the press ($\delta$, $\mu\delta$, $\tau$) at all integral elements can be quantized. At each integral element, the direction of press $\delta$ is parallel with the normal direction and the value of $\delta$ is proportional to the interference along the normal direction. The direction of friction force $\mu\delta$ is approximately along axis Z and the value of $\mu\delta$ is proportional to press $\delta$. $\tau$ represents the fiction along tangential direction, the sign and value of which are both hard to quantitatively model because $\tau$ is a kind of static friction. FPM function is derived by integrating the press at all integral elements in {A}.

$$\begin{aligned}
F_x' &= \int_{-l/2}^{l/2}\int_{\theta_l}^{\theta_u} R(\theta)\delta(\theta,s)\cos\chi(\theta)\,\mathrm{d}\theta\,\mathrm{d}s \\
F_y' &= \int_{-l/2}^{l/2}\int_{\theta_l}^{\theta_u} R(\theta)\delta(\theta,s)\sin\chi(\theta)\,\mathrm{d}\theta\,\mathrm{d}s \\
F_z' &= \int_{-l/2}^{l/2}\int_{\theta_l}^{\theta_u} R(\theta)\mu\delta(\theta,s)\,\mathrm{d}\theta\,\mathrm{d}s \\
M_x' &= \int_{-l/2}^{l/2}\int_{\theta_l}^{\theta_u} (s+\mu R(\theta))R(\theta)\delta(\theta,s)\sin\chi(\theta)\,\mathrm{d}\theta\,\mathrm{d}s \\
M_y' &= \int_{-l/2}^{l/2}\int_{\theta_l}^{\theta_u} (s+\mu R(\theta))R(\theta)\delta(\theta,s)\cos\chi(\theta)\,\mathrm{d}\theta\,\mathrm{d}s \\
M_z' &= \int_{-l/2}^{l/2}\int_{\theta_l}^{\theta_u} R^2(\theta)(\sin(\chi(\theta)-\theta)\delta(\theta,s)+\cos(\chi(\theta)-\theta)\tau(\theta,s))\,\mathrm{d}\theta\,\mathrm{d}s
\end{aligned} \tag{6}$$

where $s$ and $\theta$ are integral variables along the axis and circumference. $R(\theta)$ and $\chi(\theta)$ represent the radius and normal direction angle at integral element ($\theta, s$). $\mu$ represents the friction coefficient.

***note* 1**: As shown in (6), asymmetric and irregular geometric features make the direction of stress not parallel with the axis of {S}, thus causing the couplings between $p$ and $F$'. Specifically, a component bias of $p$ may cause more than one component biases of $F$'.

***note* 2**: When deriving FPM function, the influence of $\tau$ on $F_x$', $F_y$', and $F_z$' can be ignored because $\delta > \mu\delta \gg \tau$. But the influence of $\tau$ on $M_z$' cannot be ignored due to the amplification of the moment arm. Since $\tau$ is uncertain in real scenarios, $M_z$' in (6) is just a representation of a computational idea and cannot be modeled analytically.

## 2.2. Model-hybrid Compliance Control
### 2.2.1. Model-Based Control Modules

As shown in Fig. 1 (b), decoupling modules for couplings caused by misaligned coordinate systems are designed separately.

$$F_{dec}(t) = \mathrm{DO}(F(t)|R_S^A, t_S^A) \tag{7}$$
$$\Delta r_c(t) = \mathrm{DS}(\Delta p_c(t)|R_R^A, t_R^A) \tag{8}$$

where DO and DS are decoupling modules for output and state equations. $F_{dec}$ is the estimated $F$' in {A}. $\Delta r_c$ is the adjustment of input. $\Delta p_c$ represents the estimated adjustment of state. The concrete form of DO and DS can be referred to [12] and [42].

The couplings between components of $F_{dec}$ and $\Delta p_c$ are ignored in the model-based modules and settled by the RL agent. Each component of $F_{dec}$ only correlates to the correspondent component of $\Delta p_c$ in the model-based compliance law. The model-based compliance law is

$$\Delta p_c(t) = \mathrm{diag}(K)(F_{dec}(t) - F_{rfr}(t)) \tag{9}$$

where $K = [K_x, K_y, K_z, K_\alpha, K_\beta, K_\gamma]$. $K_x, K_y, K_z, K_\alpha, K_\beta, K_\gamma$ are compliance parameters for each state component. $F_{rfr} = [0, 0, F_{rfr,z}, 0, 0, 0]$ represents reference force/moment in {A}. $F_{rfr,z}$ is usually set to be zero when the peg and hole are with clearance fit.

### 2.2.2. RL Agent Control Module

Since model-based control modules cannot handle the couplings caused by asymmetric and irregular geometric features, and force-pose mapping is variable as the peg is inserted deeper, an RL agent is combined in the controller to improve the comprehensive control effect.

The state in Markov decision process (MDP) is defined as the output of the assembly process. The action in MDP is defined as the revision factors of compliance parameters.

$$\begin{aligned} s &= [r, F] \\ a &= [\Delta p_x, \Delta p_y, \Delta p_z, \Delta p_\alpha, \Delta p_\beta, \Delta p_\gamma] \end{aligned} \quad (10)$$

All elements of action are within the range of [$lb$, $ub$], $lb \in \mathbb{R}^6$, $ub \in \mathbb{R}^6$. Empirically, $lb$ = [-1, -1, -1, -1, -1, -1], $ub$ = [1, 1, 1, 1, 1, 1].

The reward function is

$$r = -h_z z - h_F \|[F_x, F_y, F_z]\|_2 - h_M \|[M_x, M_y, M_z]\|_2 \quad (11)$$

where $h_z$ is the coefficient for insertion speed. $h_F$ and $h_M$ are the coefficients for contact force and moment.

The adaptive compliance law combines the expert constant compliance law and the RL agent.

$$\begin{aligned} \Delta p_c(t) &= \text{diag}(K)(F_{dec}(t) - F_{rfr}(t)) \\ K &= a \circ K + K \end{aligned} \quad (12)$$

where $K = [K_x, K_y, K_z, K_\alpha, K_\beta, K_\gamma]$. $K_x, K_y, K_z, K_\alpha, K_\beta, K_\gamma$ are adaptive compliance parameters for each state component. $\circ$ represents Hadamard product. The RL agent considers the influence of other coupling force/moment components and variable force-pose mapping through adaptive revision factors. Hence, MHCC method is capable to ensure the compliance, precision and stability of the control process at the same time.

## 3. Control Reconfiguration

When constructing the controller for a certain assembly object, model-based control modules are first tuned and an RL agent is then trained. Because the model of the assembly process is tightly related to the geometric features of assembly objects, the optimal controllers for different assembly objects are different. To avoid the retuning and relearning cost of model-based modules and the RL agent, the control reconfiguration method is proposed in this section.

### 3.1. Formulation

As shown in Fig. 3, the assembly tasks, controllers of which have been constructed are treated as source domains. The assembly tasks, controllers of which are to be constructed are treated as target domains. Different geometric features of assembly objects cause the gap between source and target domains. The mission of control reconfiguration is to adapt to this gap. The controllers of assembly tasks in source domains are obtained through MHCC method, which consists of two different steps: model-based parameters tuning and RL agent training. While the controller in the target domain is obtained through the control reconfiguration method, which consists of two different steps: model-based reconfiguration and RL agent transfer.

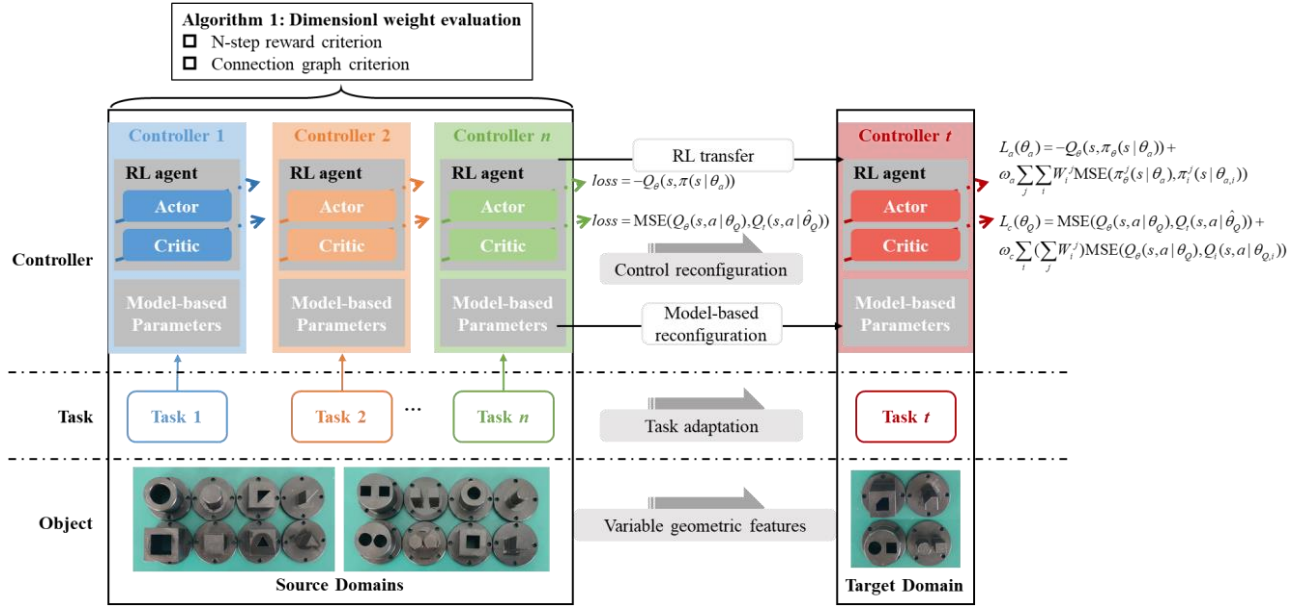

Fig. 3. Control reconfiguration method for MHCC controllers.

### 3.2. Model-Based Reconfiguration

In an MHCC controller, model-based control modules and the RL agent are coupled for an assembly task. The RL agent will be different if model-based control modules are configured differently. To reduce the gap between RL agents in source and target domains, model-based control modules should be accurately reconfigured according to the model of the assembly process.

As shown in Fig. 1 (b), when reconfiguring model-based control modules, including DS, DO, depth estimation, and compliance law, their structures are kept stable but parameters need to be reconfigured according to the geometric features of assembly objects. Since DS, DO and depth estimation are irrelevant to geometric features, we mainly focus on the reconfiguration of compliance parameters in compliance law.

According to (6), $M_z'$ cannot be modeled analytically because of the uncertainty of $\tau$. Hence, only $K_x, K_y, K_z, K_\alpha, K_\beta$ are reconfigured and $K_\gamma$ is kept the same for different assembly tasks. Here ETCL is proposed to establish an analytical connection between parameters and geometric features.

***Equivalent Theory of Compliance Law***: In a force-guided compliance controller of an assembly task, when the couplings caused by misaligned coordinate systems are well decoupled and the compliance law is in the format of proportions, critically stable compliance parameters $K_x, K_y, K_z, K_\alpha, K_\beta$ (largest parameters which ensure the stability of the entire control process) follow

$$\mathbb{E}_{p_i}(\frac{\partial F_i}{\partial p_i})\big|_{l=L} K_i = \text{const} \tag{13}$$

where $F_i$ and $p_i$ represent the component of $F'$ and $p$. $K_i$ indicates one of critically stable compliance parameters $K_x, K_y, K_z, K_\alpha, K_\beta$. Further, critically stable compliance parameters $K_x, K_y, K_z, K_\alpha, K_\beta$ follow

$$\begin{aligned} E\hat{R}LK_i s_i &= \text{const} \\ s_i &= \mathbb{E}_{p_i}(\frac{\partial F_i}{\partial p_i})\big|_{l=L} / (E\hat{R}L) \end{aligned} \tag{14}$$

where $E$ represents elasticity modulus. $\hat{R}$ represents the largest radius of the cross section. $L$ represents the depth of the hole. $s_i$ represents a shape scale, which is only determined by the shape of the cross section.

***note* 1**: According to (6), only $\partial F_i/\partial p_j$ of regular objects can be obtained analytically, while $\partial F_i/\partial p_j$ of irregular objects needs to be calculated through numerical computation.

***note* 2**: As for asymmetric assembly objects, $\partial F_i/\partial p_j$ is anisotropic in different directions. $\partial F_x/\partial d_x$ is constant if the sign of $d_x$ is kept the same. But $\partial F_x/\partial d_x$ becomes different when the sign of $d_x$ is changed. Hence, ETCL is proposed based on the expectation of $\partial F_i/\partial p_j$. In practical applications, the expectation can be calculated through $\partial F_i/\partial p_j$ at different state points. For example, we assess the expectation value by calculating $\partial F_x/\partial d_x$ at $(d_x, 0, L, 0, 0, 0)$ and $(-d_x, 0, L, 0, 0, 0)$ where the peg and hole can contact with each other.

***note* 3**: According to numerical computation results of (14), scale $s_i$ is only determined by the shape of the cross section and unrelated to other factors like $\hat{R}$ and $L$. Hence, (14) is more convenient to reconfigure model-based parameters than (13) when only the size factors of assembly objects are changed.

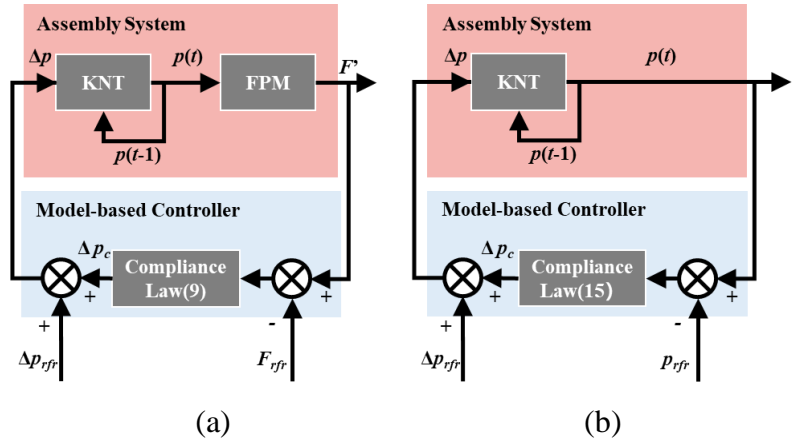

Fig. 4. Equivalent theory of compliance control in assembly tasks. (a) Simplified control block diagram assuming that decoupling modules DO and DS are accurate enough; (b) Simplified control block diagram using virtual state feedback to replace output feedback.

***Proof***: In typical assembly equipment, the couplings in state and output equations are caused by the misalignment between coordinate systems. DS and DO are designed to be corresponding decoupling modules of CS and CO. Hence, DS and DO are also determined by the misalignment between coordinate systems, which is available through insertion depth estimation and some calibration approaches. As DS and DO are essentially irrelevant to geometric features, the format of DS and DO can be directly adjusted in both source and target domains.

The model-based parameters which need to be reconfigured are proportional gains $K_x, K_y, K_z, K_\alpha, K_\beta$ in compliance law. Assuming that DS and DO are both well designed in all domains, CS and DS (CO and DO) are merged as a unit element, the model-based controller is simplified as Fig. 4 (a).

In the simplified control block diagram, KNT is irrelevant to assembly objects, while FPM relies on the geometric features of assembly objects. Here we form a virtual state feedback loop to replace the output feedback loop.

$$\Delta p_c(t) = \text{diag}(K_s)(p(t) - p_{rfr}(t)) \tag{15}$$

where $K_s = \left[ K_{s,x}, K_{s,y}, K_{s,z}, K_{s,\alpha}, K_{s,\beta}, K_{s,\gamma} \right]$. $K_{s,x}, K_{s,y}, K_{s,z}, K_{s,\alpha}, K_{s,\beta}, K_{s,\gamma}$ are state feedback gains for state components.

As shown in Fig. 4 (b), the compliance law in state feedback is in the same format as output feedback. Since KNT is irrelevant to assembly objects, the state feedback gains are not affected by geometric features.

When tuning compliance parameters, we try to eliminate the state bias as rapidly as possible under the premise of ensuring stability in the entire control process. The biggest concern for the stability occurs at the end stage of the assembly process when the peg is totally inserted into the hole. Hence, $K$ is always tuned to stabilize the assembly at this stage. Then the effect of state feedback compliance law (15) is equal to the combination of output feedback compliance law (9) and FPM at the end stage. Because we ignore the couplings within FPM in the model-based controller, FPM is linearized as

$$F_i = \mathbb{E}_{p_i}(\frac{\partial F_i}{\partial p_i})|_{l=L} \, p_i = E\hat{R}Ls_i p_i \tag{16}$$

According to (9), (15), and (16), $K_{s,i} = K_i E\hat{R}Ls_i$. As $K_{s,i}$ always remains the same for all assembly objects with different geometric features, $K_i E\hat{R}Ls_i$ is kept constant for different assembly objects.

### 3.3. RL Agent Transfer
### 3.3.1. Weighted Dimensional Policy Distillation

This paper proposes WDPD method, which is an improved RL transfer method based on [41]. The similarities between domains are the foundation for transferring knowledge. The agent in the target domain is always desired to learn more detailed knowledge from more similar source domains. Hence, it is necessary to evaluate the similarities between source and target domains before transferring. Moreover, if the similarity evaluation can be implemented in every single dimension of policies, the agent in the target domain can identify the applicability of sub-policies in source domains more accurately. The main innovation of WDPD method lies in similarity weight and dimensional transfer mechanisms.

In this paper, the similarity is evaluated through N-step reward and the connection graph matrix [43].

$$Sim_i^j = N_i^j + \upsilon G_i(j,:)G_\theta(j,:)^T \tag{17}$$

where $Sim_i^j$ represents the similarity between sub-policy $j$ of source domain $i$ and the target domain. $N_i^j$ represents N-step reward. $G_i$ and $G_\theta$ represent the connection graph matrices in source domain $i$ and the target domain. $G_i(j,:)$ is the row vector $j$ of $G_i$, which describes the influence of action $j$ on all state components in source domain $i$. $\upsilon$ represents the coefficient to balance the magnitudes of N-step reward and connection graph matrix criteria. N-step reward and connection graph matrix are defined as

$$N_i^j = \mathbb{E}_{\pi_i^j}(\sum_{t=0}^{n-1} r_t) \tag{18}$$

where $n$ is the dimension of state space.

$$G_i = \left[ G_{u,v} = \frac{\phi(^u a, ^v s)}{\left\| [\phi(^1 a, ^v s), \phi(^2 a, ^v s), ..., \phi(^m a, ^v s)] \right\|_2} \right] \in \mathbb{R}^{m \times n} \tag{19}$$

where $\phi(^u a, ^v s)$ indicates continuous space serialized Shapley value in [43], $^u a$ and $^v s$ represent the component of action $a$ and the component of state $s$. $m$ is the dimension of action space.

The similarity is described from two aspects: N-step reward and connection graph matrix. A higher N-step reward value indicates that the sub-policy in the source domain is more suitable to the target domain. Connection graph matrix describes the relationship between action components and state components in a domain. A larger $G_i(j,:)G_\theta(j,:)^T$ shows that action component $^j a$ has a more similar influence on state components in both source domain $i$ and the target domain. When source domain $i$ is the same as the target domain, $G_i(j,:)G_\theta(j,:)^T = 1$. The algorithm of dimensional similarity evaluation is shown in Algorithm 1.

---

**Algorithm 1**: Dimensional similarity evaluation

**Input**: Sampling times *ST*, Coefficient $\upsilon$

**Output**: All similarities $Sim_i^j$

for each source domain $i$ and target domain:
  for each sub-policy $j$:
    for $st = 1, ST$:
      Random sample $s_0$ in target domain.
      for $t = 0, n-1$:
        Activate sub-policy $j$ $a_t = [0, ..., \pi_i^j(s_t), ..., 0]$, observe $s_{t+1}$ and $r_t$.
        Take action $a_t = [0, ..., 0]$, observe $s'_{t+1}$.
  Calculate $N_i^j$ according to (18).
  Calculate $\phi(^j a, ^v s), v = 1, 2, ..., n$.
  Calculate $G_i$ or $G_\theta$ according to (19).
  Evaluate similarity according to (17).

WDPD method is proposed to drive the agent in the target domain to learn dimensional knowledge from all sub-policies of all source domains with similarity weight. Based on the similarity evaluation, dimensional similarity weight is designed as

$$W_i^j = \frac{Sim_i^j - \min_j Sim_i^j}{\sum_j \left(Sim_i^j - \min_j Sim_i^j\right)} \tag{20}$$

According to [40], the objective function which is to be maximized in the target domain includes two parts: the part of encouraging exploring and that of learning from existing knowledge. The objective function of WDPD method is

$$J(\theta) = E_{(s_t, a_t) \sim \pi_\theta} \left[\sum_{t \geq 0} \gamma^t r_t + \omega \sum_j \sum_i W_i^j \mathcal{H}^\times(\pi_i^j | \pi_\theta^j)\right] \tag{21}$$

where $\theta$ represents the parameters of the agent in the target domain. $\gamma \in [0,1)$ represents discount rate. $\mathcal{H}^\times()$ demotes Shannon's cross entropy, which describes the divergence of two distributions. $\omega$ represents the scale to balance two parts of objectives.

### 3.3.2. Setup and Implementation

Actor-critic RL algorithms are preferred in practical applications because of the advantages of high convergence speed and stability. The agent in the assembly controller is obtained through DDPG, a typical off-policy deterministic actor-critic RL algorithm. Here we implement WDPD method in DDPG as an example to accomplish RL agent transfer.

In actor-critic algorithms, actor and critic are distilled at the same time to achieve the objective (21). Since DDPG offers a deterministic policy and is updated with off-policy mechanism, Shannon's cross entropy can be simplified as MSE function. The loss functions of actor and critic in WDPD method are

$$L_a(\theta_a) = -Q_\theta(s, \pi_\theta(s|\theta_a)) + \omega_a \sum_j \sum_i W_i^j \text{MSE}(\pi_\theta^j(s|\theta_a), \pi_i^j(s|\theta_{a,i}))$$

$$L_c(\theta_Q) = \text{MSE}(Q_\theta(s,a|\theta_Q), Q_t(s,a|\hat{\theta}_Q)) + \omega_c \sum_i (\sum_j W_i^j) \text{MSE}(Q_\theta(s,a|\theta_Q), Q_i(s,a|\theta_{Q,i})) \tag{22}$$

where $\pi_\theta$, $Q_\theta$, and $Q_t$ indicates actor, critic, and target critic in the target domain. $\theta_a$, $\theta_Q$, and $\hat{\theta}_Q$ indicates their parameters. $\pi_i$ and $Q_i$ indicates actor and critic in source domain $i$. $\theta_{a,i}$ and $\theta_{Q,i}$ indicates their parameters. Superscript $j$ represents sub-policy or action component of a policy in a domain. $\omega_a$ and $\omega_c$ indicates the scale to balance two parts of objectives.

## 4. Experiment Verifications

The primary target of experiments is to verify the effectiveness of the proposed control reconfiguration method. During the reconfiguration process, we first reconfigure the model-based parameters according to ETCL. Then the similarities between different assembly tasks are evaluated based on the reconfigured parameters. Finally, the RL agent is transferred with WDPD method and the entire controller is completed for the new task.

We verify the control and reconfiguration methods in the following experimental conditions. The experiment equipment includes an assembly robot, a force-moment sensor, and groups of assembly objects with different geometric features. As shown in Fig. 5 (a), the force-moment sensor is fixed at the end of the robot. The peg is installed below the force-moment sensor and the hole is fixed on the table. The axes of the robot, force-moment sensor, and peg are collinear, which means that the axes Z of {A}, {S}, and {R} are collinear. The force-moment sensor provides six-dimensional force/moment information $F$ and the robot provides six-dimensional Cartesian pose $r$. As shown in Fig. 5 (b), we select two groups of assembly objects with the same physical features but different geometric features to verify the control reconfiguration method.

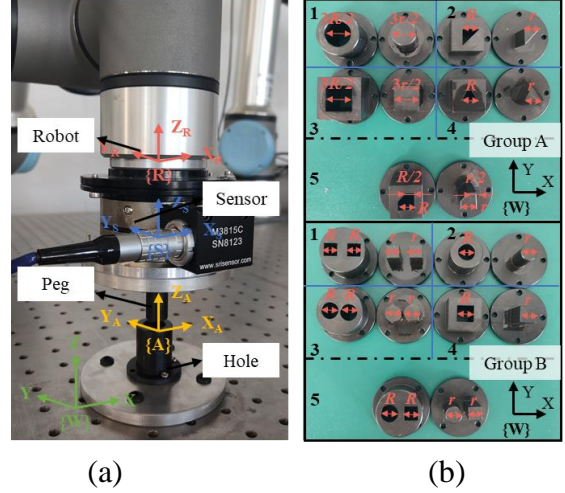

(a)      (b)

Fig. 5. Experiment equipment. (a) Robotic assembly system; (b) Assembly objects. The indicated dimensions in groups A and B are: $R = 10$ mm, $r = 9.9$ mm.

### 4.1. Model-Based Reconfiguration

Table 1. Reconfiguration of model-based parameters for objects with different geometric features.

| | Label | 1 | 2 | 3 | 4 | 5 |
|---|---|---|---|---|---|---|
| | $\hat{R}$ (mm) | 7.5 | 7.45 | 10.61 | 5.77 | 7.14 |
| | $L$ (mm) | 30 | 20 | 30 | 20 | 20 |
| Group A | $s_i / s_{i,c}$* | 1.12, 1.12, 1.41, 1.12, 1.12 | 0.81, 0.81, 0.95, 0.81, 0.81 | 1, 1, 1, 1, 1 | 1.07, 1.07, 1.23, 1.07, 1.07 | 0.92, 0.92, 0.99, 0.92, 0.92 |
| | Method | Reconfigured | Reconfigured | Tuned | Reconfigured | Reconfigured |
| | $K_x, K_y, K_z$ (e-5 m/N) | 4.29, 4.29, 2.16e-2 | 8.96, 8.96, 4.85e-2 | 3.39, 3.39, 2.16e-2 | 8.75, 8.75, 4.85e-2 | 8.19, 8.19, 4.85e-2 |
| | $K_\alpha, K_\beta, K_\gamma$ (e-2 rad/Nm) | 4.21, 4.21, 5.55* | 8.81, 8.81, 5.55* | 3.33, 3.33, 5.55* | 8.61, 8.61, 5.55* | 8.05, 8.05, 5.55* |
| | Label | 1 | 2 | 3 | 4 | 5 |
| | $\hat{R}$ (mm) | 13.06 | 5 | 12.07 | 7.07 | 13.06 |
| | $L$ (mm) | 20 | 20 | 20 | 20 | 20 |
| Group B | $s_i / s_{i,c}$* | 1.08, 1.08, 1.08, 1.08, 1.08 | 1.12, 1.12, 1.41, 1.12, 1.12 | 0.92, 0.92, 1.17, 0.92, 0.92 | 1, 1, 1, 1, 1 | 0.97, 0.97, 1.08, 0.97, 0.97 |
| | Method | Reconfigured | Reconfigured | Reconfigured | Tuned | Reconfigured |
| | $K_x, K_y, K_z$ (e-5 m/N) | 3.82, 3.82, 2.43e-2 | 9.66, 9.66, 4.85e-2 | 4.86, 4.86, 2.43e-2 | 7.63, 7.63, 4.85e-2 | 4.26, 4.26, 2.43e-2 |
| | $K_\alpha, K_\beta, K_\gamma$ (e-2 rad/Nm) | 3.75, 3.75, 5.55* | 9.48, 9.48, 5.55* | 4.77, 4.77, 5.55* | 7.49, 7.49, 5.55* | 4.18, 4.18, 5.55* |

\* $s_{i,c}$ represents the shape scale of cuboid object whose compliance parameters are tuned manually in each group. Subscript $i$ represents the component label, $i = x, y, z, \alpha, \beta$. $K_\gamma$ remains the same across different assembly tasks and will not be manually tuned again unless it causes control performance degradation.

If the geometric features of an object are known, compliance parameters can be reconfigured to a new object according to ETCL. In each group, the compliance parameters of cuboid object (label 3 in group A and label 4 in group B) are tuned manually while the compliance parameters of the other objects are reconfigured according to the change of geometric features. Geometric features and compliance parameters are shown in Table. 1. As shown in Fig. 5 (b), all assembly objects are shaped as columns of equal cross section. The axis is the combination of gravity centers of all cross section. $\hat{R}$ represents the largest distance from a point on the edge of a cross section to the gravity center of the cross section. $L$ represents the depth of the hole, also the depth of the peg. To simplify the format, we record a relative $s_i$ in Table 1.

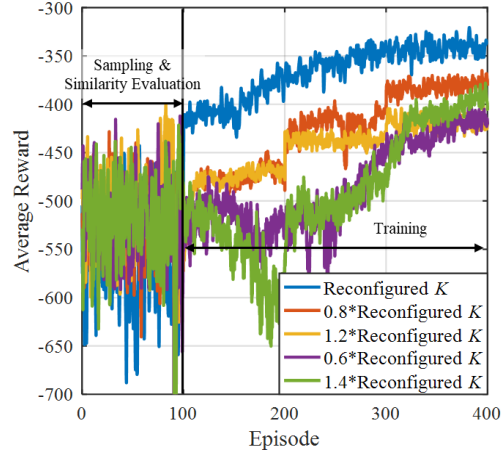

Fig. 6. Average reward curves of label 5 with different compliance $K$ when transferring only from label 3 in group A. Reconfigured $K$ is shown in Table 1.

To verify the effectiveness of ETCL on model-based reconfiguration, we transfer the agent of label 3 to train the agent of label 5 in group A but set different compliance parameters $K$ for label 5 before training. During transferring, $\omega_a$ is set to be 2 and $\omega_c$ is set to be 5.

It is found that reconfigured $K$ through ETCL has an obvious positive effect on the transfer process. However, when compliance parameters violate ETCL, the transferred knowledge in label 3 may mislead the agent of label 5. Especially when the deviations of compliance parameters become larger, the effect of RL agent transfer becomes lower.

The results demonstrate that ETCL helps to narrow the gap between source and target domains, thus increasing the positive effect of knowledge transfer from the source domains to the target domain.

### 4.2. RL Agent Transfer

WDPD method argues that the agent in the target domain should transfer dimensional knowledge from more similar source domains. To show the effect of similarity evaluation and WDPD algorithm, we train RL agents on the first four objects separately and transfer the trained agents to the last object in each group. Before evaluating similarity and training, all compliance parameters are set as shown in Table 1.

During similarity evaluation, $\upsilon$ is set to be 100 according to the magnitude difference between N-step reward and connection graph criteria. According to Algorithm 1, we implement each sub-policy of each source domain in the target domain separately. We calculate the transfer weight through (18) according to all $Sim_i^j$. The dimensional transfer weight is shown in Fig. 7.

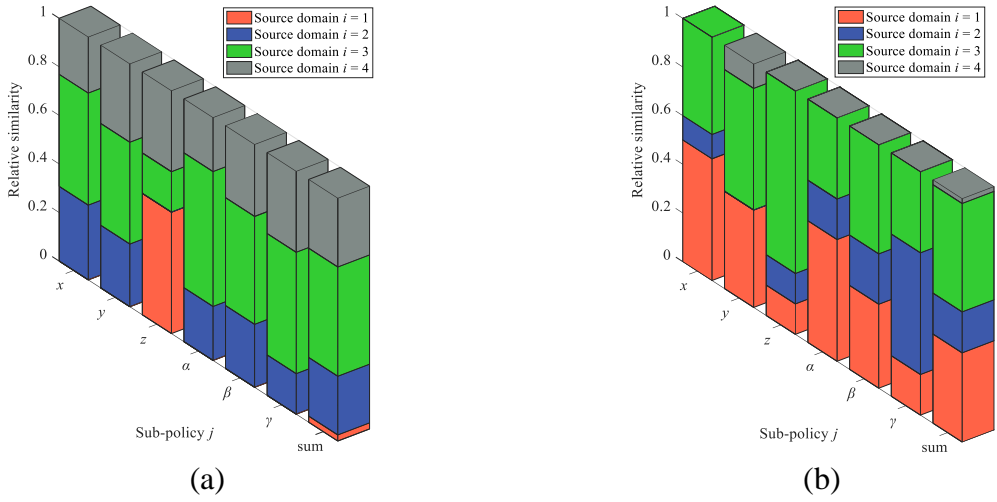

Fig. 7. Dimensional similarity weight between sub-policies in the source domains and sub-policies in the target domain. (a) Group A; (b) Group B.

The similarity depends on the geometric features of these objects to some extent. As shown in Fig. 7 (a), the assembly object shaped like a pentagonal prism is most similar to that of a cuboid, but least similar to that of a cylinder. As shown in Fig. 7 (b), the assembly object composed of a cylinder and a cuboid is more similar to dual cylinders or cuboids than a single cylinder or cuboid. Besides, because similarity evaluation is implemented after the reconfiguration of model-based parameters, the similarity is related to not only the geometric features of objects but also the accuracy of force-pose mapping. If force-pose mapping is not accurate enough and the model-based parameters are not reconfigured well, the similarity may be still low even though two assembly objects have the same geometry.

WDPD method focuses on how to utilize the knowledge in source domains more efficiently to accelerate the training process in the target domain. Similarity weight and dimensional transfer are two main core mechanisms. As shown in Fig. 8, the average reward curves obtained by different methods are compared. Four baselines for WDPD method are selected: the directly trained agent, the agent transferred equally from all source domains, the agent transferred only from the most similar source domain (all sub-policies in this domain), and the agent transferred only from the least similar source domain (all sub-policies in this domain).

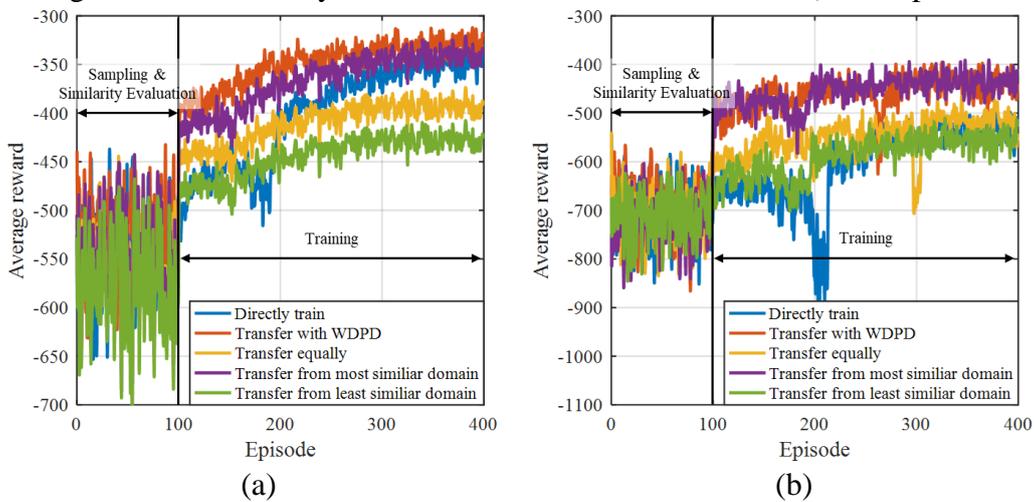

Fig. 8. Average reward curves of label 5 with different methods. (a) Group A; (b) Group B.

We mainly focus on the average reward under different learning costs during the training process, which quantitatively characterize the average transfer performance. The first 100 episodes are used to sample data in replay buffer as well as evaluate similarity at the same time. As shown in Fig. 8 (a), WDPD method generates the highest increase in the reward curve once the agent starts to update. Besides, WDPD method achieves the largest final reward at the end stage of the transfer process. Only transferring from the most similar source domain also has a positive effect, which is slightly weaker than WDPD method. Meanwhile, equally transferring and only transferring from the least similar source domain can only perform better than directly training method at the beginning of the training process, but the dissimilar knowledge limits the exploration of the agent in the target domain, which damages the final reward near the end stage. As shown in Fig. 8 (b), WDPD method still performs best in group B. In conclusion, WDPD method is more effective to assist the agent in the target domain to achieve higher reward in less training time than other baseline methods.

The results demonstrate that both similarity weight and dimensional transfer are effective mechanisms to extract and utilize more valuable single-dimensional knowledge from source domains. WDPD method can accelerate the training process of the target domain more efficiently.

### 4.3. Testing Process

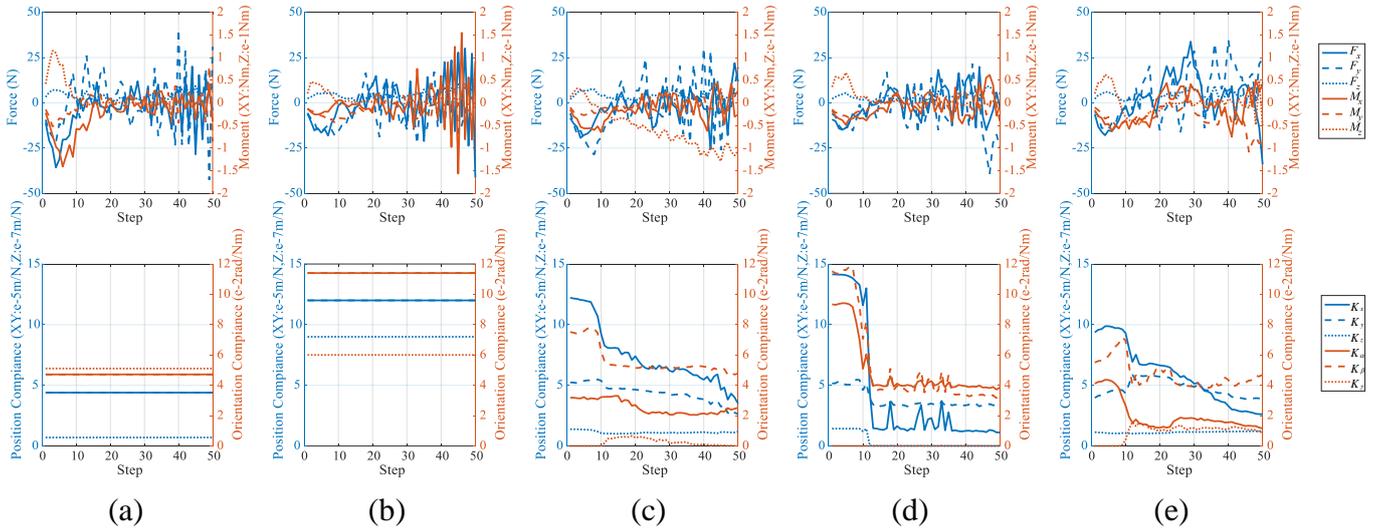

Fig. 9. Force/moment and compliance parameters curves of label 5 in group A during the testing process. (a) Constant compliance controller with relatively small compliance parameters; (b) Constant compliance controller with relatively large compliance parameters; (c) Directly trained controller; (d) Reconfigured controller with WDPD method; (e) Reconfigured controller with equal weighted policy distillation.

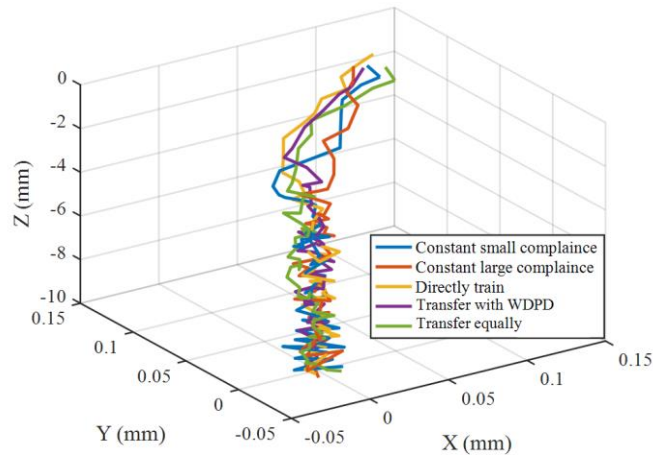

Fig. 10. Assembly trajectories of robot TCP in {W}.

We compare the performance of WDPD method reconfigured controller with several baseline controllers, including constant compliance controllers, a directly trained controller, and a reconfigured controller with equal weighted policy distillation. All the model-based modules in the directly trained controller, reconfigured controller with equal weighted policy distillation, and reconfigured controller with WDPD follow ETCL. The force/moment curves and compliance parameters curves in each controller are shown in Fig. 9. The assembly position trajectories of robot TCP are shown in Fig. 10.

As shown in Fig. 9 (a) and (b), if the constant controller is endowed with relatively small compliance parameters, force/moment curves will first reach a larger peak and then decrease more steadily. If the constant controller is endowed with relatively large compliance parameters, the peak will be smaller but the control process will be less stable. As shown in Fig. 9 (c) - (e), RL agents adjust compliance parameters adaptively. Hence, the rapidity and stability of the controller can be guaranteed at the same time. The reconfigured controller with WDPD method provides the smallest average force/moment, which is consistent with the average reward curves in the training process (Fig. 8 (a)).

The results demonstrate that the proposed control reconfiguration method (model-based reconfiguration based on ETCL and RL agent transfer based on WDPD) can achieve a better control effect than other controllers in the same limited time.

## 5. Conclusions

This paper employs MHCC method to construct the controller for assembling an object with certain geometric features. To save the relearning cost for new assembly objects with different geometric features, this paper further proposes a control reconfiguration method to transfer the existing control knowledge to new assembly tasks.

The main contributions of this paper are summarized as follows. Firstly, this paper proposes a control reconfiguration method for MHCC controllers, which is composed of model-based reconfiguration and RL agent transfer in sequence. The reconfiguration method can provide better control performance with less training cost. Secondly, this paper proposes ETCL to accomplish model-based reconfiguration which can relate compliance parameters with geometric features directly. Model-based reconfiguration greatly reduces the gap between the agents in source and target domains, which provides a foundation for RL agent transfer. Last but not least, this paper proposes WDPD method to transfer single-dimensional prior knowledge from the source domains to the target domain. WDPD method can extract and utilize more valuable single-dimensional knowledge and then accelerate the transfer process more efficiently.

The proposed control reconfiguration method provides a paradigm for solving assembly control of objects with different geometric features, which has great practical significance for flexible and diversified assembly scenarios, like electronic, communication, and automotive fields.

In the future, how to transfer knowledge from simulation scenarios to practical scenarios in contact-rich assembly tasks is an attractive research direction.